\documentclass[letterpaper]{article} 
\usepackage[preprint]{aaai2027} 
\usepackage[hyphens]{url} 
\usepackage{graphicx} 
\usepackage{natbib} 
\usepackage{caption} 
\usepackage{amsmath}
\usepackage{amssymb}
\usepackage{booktabs}
\usepackage{multirow}
\usepackage{array}
\usepackage{tabularx}
\usepackage{fvextra}
\usepackage{seqsplit}

\newcommand{\ours}{\textsc{Oracle-\allowbreak OPD}}
\newcommand{\bench}{\textsc{Context\allowbreak Pollute-\allowbreak Bench}}
\newcommand{\call}{\textsc{Call}}
\newcommand{\answer}{\textsc{Answer}}
\newcommand{\best}[1]{\textbf{#1}}

\newcommand{\legacyop}{legacy-param+success}

\DefineVerbatimEnvironment{CodeBlock}{Verbatim}{
  fontsize=\footnotesize,
  breaklines=true,
  breakanywhere=true,
  breaksymbolleft={},
  frame=single,
  framesep=2pt,
  xleftmargin=0pt,
  xrightmargin=0pt
}

\newcolumntype{Y}{>{\raggedright\arraybackslash}X}
\newcommand{\codeid}[1]{\texttt{\seqsplit{#1}}}

\title{When History Lies: Evaluating and Improving Tool Use\\
under Misleading Multi-Turn Histories}

\author{
Xiaoqing Wu\textsuperscript{\rm 1},
Xingyu Fan\textsuperscript{\rm 1},
Feifei Li\textsuperscript{\rm 1},
Wenhui Que\textsuperscript{\rm 1}
}

\affiliations{
\textsuperscript{\rm 1}WeChat, Tencent Inc., Beijing, China\\
\{xiaoqingwwu,fanxfan,niyali,victorque\}@tencent.com
}

\begin{document}
\maketitle

\begin{abstract}
Tool-calling agents infer task state from accumulated dialogue and tool traces.
In persistent interactions, however, historical traces may remain structurally
valid and semantically plausible after they cease to be authoritative for the
current request. We show that such history can \emph{hijack a policy the model
already possesses}: on Qwen3-1.7B, pollution flips $32.1\%$ of decisions that
are correct under the original trajectory and frequently induces reuse of
corrupted entities or interface conventions. We introduce \bench, a paired
benchmark with synchronized Original, Polluted, and Oracle State views that
preserve the system policy, current tools, latest request, and gold next action.
Eleven gold-preserving interventions isolate failures in decision state, entity
binding, and interface execution across complete calls and non-call decisions.
We further propose \ours, which transfers an Oracle-conditioned teacher policy
to a student observing only polluted history through soft supervision on
student-generated prefixes. On Qwen3-1.7B, \ours{} achieves $87.0\%$ Balanced
Tool-Use Accuracy, outperforming Gold-SFT ($66.3\%$), Oracle sequence
distillation ($82.3\%$), and off-policy token distillation ($85.0\%$).
The method scales consistently: an 8B teacher raises the same compact 1.7B
student to $91.9\%$, while an 8B student reaches $93.0\%$. The resulting
policies further transfer to clean histories, unseen functions, independently
regenerated evaluation contexts, external tool-use benchmarks, and noisy
multi-hop question answering. These results establish history reliability as a
distinct tool-use bottleneck and demonstrate reliable-state policy transfer as
an effective and scalable solution.
\end{abstract}

\section{Introduction}

Tool-using language models interleave reasoning with actions, learn tool-use
behavior through prompting or self-supervision, and operate over large API
collections \cite{yao2023react,schick2023toolformer,patil2024gorilla}.
In realistic deployments, however, they rarely act on a pristine prompt.
Instead, they consume an accumulating interaction record in which users revise
requests, tools return partial or failed observations, and earlier turns leave
behind identifiers, arguments, and state claims. Such information may remain
syntactically valid and semantically plausible after it has stopped governing
the current action.

\begin{figure*}[t]
\centering
\IfFileExists{fig1_new.pdf}{
  \includegraphics[
    width=0.98\textwidth
  ]{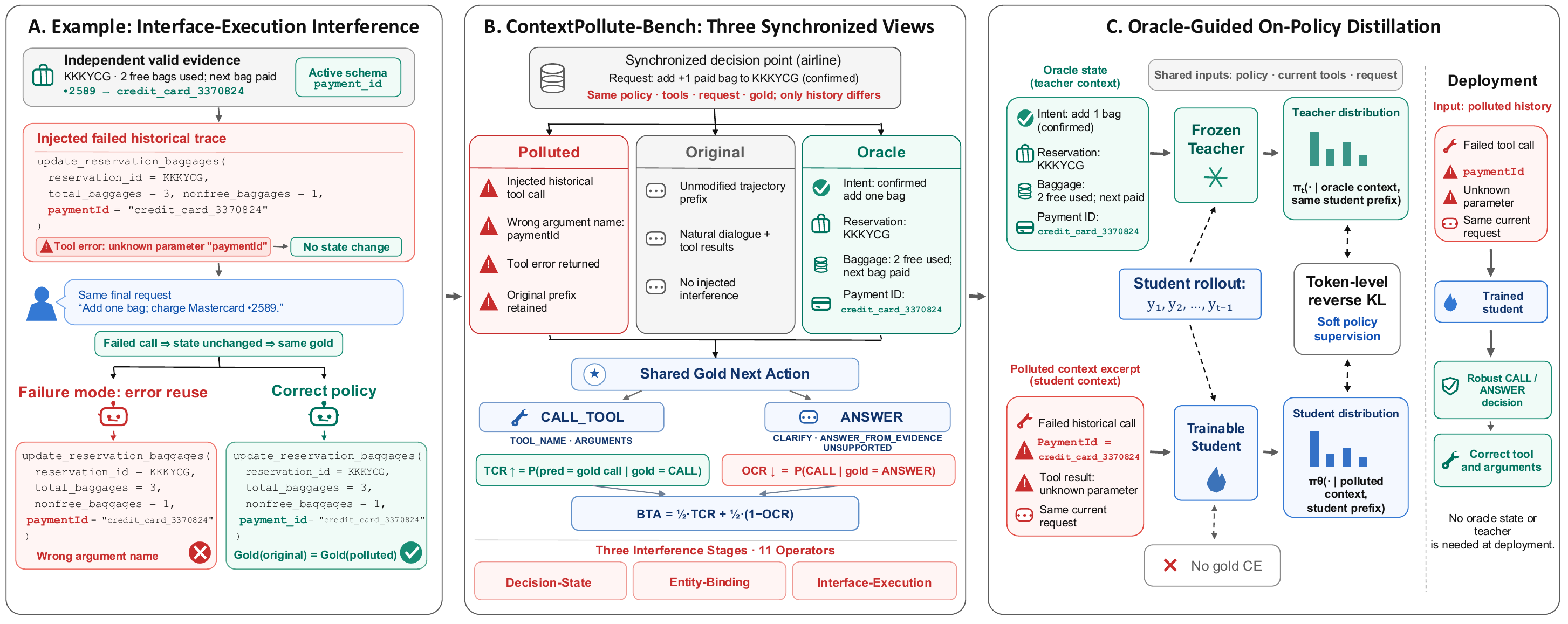}
}{
  \fbox{\parbox[c][3.2cm][c]{0.94\textwidth}{\centering
  \textbf{Missing file: fig1\_new.pdf}}}
}
\caption{Overview of the problem, benchmark, and method. (A) An example in which a failed historical interface call precedes the current request while leaving the task state and gold action unchanged. (B) The synchronized Polluted, Original, and Oracle State views in \bench{}, together with the evaluated \call{} and \answer{} decisions. (C) \ours{} training with polluted-context student rollouts and Oracle-conditioned teacher supervision on the same prefixes. Deployment uses only the trained student and ordinary interaction history.}
\label{fig:overview}
\end{figure*}

Figure~\ref{fig:overview}A illustrates this failure. An airline assistant has
the evidence needed to update reservation \texttt{KKKYCG}, but the history also
contains a failed tool call using the invalid argument name
\texttt{paymentId}. Because that failure does not change the reservation state,
the correct next action is identical with or without the trace. A robust model
should follow the current schema and emit \texttt{payment\_id}; instead, it may
imitate the misleading historical precedent. Similar failures arise from
traces involving another entity, unsupported completion claims, or tools that
are no longer available.

Existing tool-use benchmarks span executable API suites, large-scale API
collections, function selection and argument generation, stateful interaction,
and when-to-call behavior \cite{li-etal-2023-api,qin2024toolllm,patil2025the,liu2025toolace,prabhakar2026apigenmt,lu-etal-2025-toolsandbox,yao2025taubench,ross-etal-2025-when2call}. Recent robustness work studies natural query
variations and changes to the available toolkit
\cite{rabinovich-anaby-tavor-2025-robustness}, while ReflecTool-\allowbreak Bench
evaluates whether models detect and repair prior tool-use mistakes in
multi-turn dialogues \cite{liu-etal-2026-llms}. These settings do not isolate
whether a plausible but non-authoritative historical trace redirects the model
when the current request, available tools, and gold action remain fixed.
Consequently, aggregate errors cannot distinguish a model that lacks the
required policy from one that possesses it under reliable context but is
redirected by misleading history.

We study this failure mode through a controlled intervention. For each decision
point, we construct three synchronized views: the untouched Original
trajectory, a Polluted trajectory containing an inserted misleading trace, and
an Oracle State view containing only reliable task state. The system policy,
tool set, latest request, and gold next action remain identical across views.
This paired design directly tests whether misleading history flips a decision
that the model can otherwise execute correctly.

The effect is substantial. On Qwen3-1.7B
\cite{yang2025qwen3technicalreport}, pollution flips $32.14\%$ of examples
that are correct under Original history, affecting both \call{} and \answer{}
decisions; direct reuse of injected entities and interface forms further
identifies the misleading trace as the cause. We call this phenomenon
history-induced policy hijacking.

To measure and mitigate it, we introduce \bench{}
(Figure~\ref{fig:overview}B), which pairs synchronized Original, Polluted, and
Oracle State views under eleven gold-preserving interventions. We further
propose Oracle-guided On-Policy Distillation
(\ours{}; Figure~\ref{fig:overview}C), where a frozen Oracle-conditioned
teacher supervises prefixes generated by a student observing only Polluted
context. Deployment uses only the student and ordinary history. \ours{} reaches
$0.8703$ BTA on Qwen3-1.7B, and an 8B teacher raises the same compact student
to $0.9193$.

Our contributions are:
\begin{itemize}
    \item \textbf{Controlled diagnosis.}
    \bench{} isolates history-induced policy hijacking through synchronized,
    gold-preserving Original, Polluted, and Oracle State views over complete
    calls and non-call decisions.
    \item \textbf{Reliable-state policy transfer.}
    \ours{} transfers an Oracle-conditioned teacher policy onto
    polluted-context student rollouts without requiring Oracle State or a
    teacher at deployment.
\item \textbf{Broad validation.}
Paired attribution, operator-level analysis, clean-view and cross-generator
transfer, unseen functions, model scaling, and external tool-use and noisy
multi-hop QA evaluations establish the mechanism and breadth of the resulting
robustness.
\end{itemize}

\section{Related Work}

\paragraph{Tool-use evaluation and robustness.}
Tool-using language models have been studied through reasoning--action
frameworks, self-supervised tool use, and large-scale API access
\cite{yao2023react,schick2023toolformer,patil2024gorilla}. Existing benchmarks
cover executable APIs and training resources
\cite{li-etal-2023-api,qin2024toolllm}, function and argument correctness
\cite{patil2025the,liu2025toolace}, stateful interaction
\cite{lu-etal-2025-toolsandbox,yao2025taubench}, verifiable multi-turn
trajectories \cite{prabhakar2026apigenmt}, and when-to-call behavior
\cite{ross-etal-2025-when2call}. Robustness work varies queries or available
tools \cite{rabinovich-anaby-tavor-2025-robustness}; ReflecTool-\allowbreak Bench evaluates
repair of prior mistakes \cite{liu-etal-2026-llms}; and InjecAgent and
AgentDojo study malicious instructions in tool content
\cite{zhan-etal-2024-injecagent,debenedetti2024agentdojo}. In contrast,
\bench{} keeps the current request, tools, and gold action fixed and intervenes
only on non-authoritative history, isolating policy preservation rather than
task variation, error repair, or instruction injection.

\paragraph{Context and policy distillation.}
Multi-turn presentation can alter model behavior even when task information is
preserved \cite{laban2026llms,chen2026conversationllmsteachclose}. Knowledge
and sequence-level distillation transfer teacher distributions or decoded
outputs
\cite{hinton2015distillingknowledgeneuralnetwork,kim2016sequencelevelknowledgedistillation},
while reverse-KL and on-policy distillation supervise generative students on
their visited states \cite{gu2024minillm,ICLR2024_5be69a58}. Learning from
privileged information and context distillation study supervision available
during training but absent at deployment
\cite{VAPNIK2009544,lopezpaz2016unifyingdistillationprivilegedinformation,snell2022learningdistillingcontext,ye2026onpolicycontextdistillationlanguage,penaloza2026privileged}.
\ours{} specializes these principles to reliability-asymmetric tool use, with
an Oracle-conditioned teacher supervising a student that operates on
potentially misleading histories.
\section{Methodology}

\subsection{\bench}

\subsubsection{Task and synchronized views.}

Given system policy $p$, available tools $\mathcal{T}$, history $h_t$, and
latest request $u_t$, the model predicts
\begin{equation}
a_t \in
\left\{
\operatorname{ANSWER}(y),
\operatorname{CALL}(f,\mathbf{z})
\right\}.
\end{equation}
A \call-required prediction is correct only if it invokes the gold tool with
the complete gold argument object. An \answer-required prediction should
clarify missing information, answer from verified evidence, or recognize that
the current tools cannot complete the request.

Each benchmark item is a synchronized tuple
\begin{equation}
\mathcal{D}_i =
\left(
x_i^{\mathrm{orig}},
x_i^{\mathrm{poll}},
x_i^{\mathrm{oracle}},
a_i^\star
\right),
\end{equation}
whose views share the system policy, tools, latest request, and gold action.
Original is the untouched source prefix.
Polluted inserts a structurally valid but non-authoritative trace
before the unchanged request, while preserving
\begin{equation}
a^\star(x_i^{\mathrm{orig}})
=
a^\star(x_i^{\mathrm{poll}})
=
a^\star(x_i^{\mathrm{oracle}}).
\label{eq:gold_preserving}
\end{equation}
Oracle State contains only reliable, decision-relevant intent, slots,
observations, policy, and evidence status. It preserves the information needed
for the decision without exposing the gold tool name.
\subsubsection{Construction and validation.}

We construct \bench{} from the airline and retail domains of APIGen-MT
\cite{prabhakar2026apigenmt}. A \call{} decision is retained when the next source message is a valid call to
an available tool, which directly provides the gold tool and arguments.
\answer{} examples are user turns followed by substantive non-tool responses
and cover \textsc{Clarify}, \textsc{Answer-from-Evidence}, and
\textsc{Unsupported} decisions.

DeepSeek-\allowbreak V4-\allowbreak Flash \cite{deepseekai2026deepseekv4} generates the inserted pollution trace and Oracle
State, but not the source gold calls or answers. Eleven operators span
Decision-State, Entity-Binding, and Interface-\allowbreak Execution interference, including
unsupported completion claims, irrelevant reads, decommissioned tools,
competing entities, and wrong, missing, extra, obsolete, reformatted, or
unit-shifted arguments.

Deterministic validators preserve source-message order, keep the latest request
final, pair tool calls with results, instantiate the intended distractor, and
verify that the gold action is unchanged. Oracle States are rejected if they
expose the gold tool or omit decision-critical state. Complete operator
definitions, generation prompts, and deterministic validation details are
provided in Appendix~A.

\subsubsection{Splits, audit, and metrics.}

All examples derived from the same source trajectory remain in the same split,
preventing overlap across train, validation, and test sets. Test-Indist uses
the same gold-tool inventory as training, whereas OOD-600 contains 600
\call-required examples whose gold actions invoke six tools never observed as
gold tools during training. Table~\ref{tab:data} summarizes the resulting
splits.

\begin{table}[b]
\centering
\small
\setlength{\tabcolsep}{4pt}
\begin{tabular}{@{}lrrrr@{}}
\toprule
Split & Total & \call{} & \answer{} & Gold tools\\
\midrule
Train       & 9,901 & 5,828 & 4,073 & 14\\
Validation  & 1,121 &   846 &   275 & 13\\
Test-Indist & 2,375 & 1,651 &   724 & 13\\
OOD-600     &   600 &   600 &     0 & 6 held out\\
\bottomrule
\end{tabular}
\caption{Benchmark split statistics. The table reports the total number of
examples, the numbers of \call{} and \answer{} examples, and the gold-tool
inventory in each split.}
\label{tab:data}
\end{table}

We validate benchmark quality with an adjudicated 200-example audit and a
stratified 1,000-example cross-audit spanning data splits, decision types, and
interference operators. The first yields $97.0\%$ gold correctness, $99.0\%$
Oracle sufficiency, and $98.0\%$ annotator agreement, with no literal gold-tool
leakage in 100 audited \call{} examples. In the larger audit, humans accept
$93.7\%$ of examples jointly across gold validity, Oracle quality, and
pollution naturalness; Claude-Opus-4.7 accepts $91.6\%$, with $92.5\%$
agreement on joint pass/fail decisions. Full protocols and per-criterion
results are provided in Appendix~B.

Let $d^\star$ and $\hat d$ denote the gold and predicted decisions. For gold
\call{} examples, $(f^\star,\mathbf{z}^\star)$ and
$(\hat f,\hat{\mathbf{z}})$ denote the gold and predicted tool calls:
\begin{equation}
\mathrm{TCR}
=
P\!\left(
\hat d=\operatorname{CALL},
\hat f=f^\star,
\hat{\mathbf z}=\mathbf z^\star
\mid d^\star=\operatorname{CALL}
\right),
\end{equation}

\begin{equation}
\mathrm{OCR}
=
P\!\left(
\hat d=\operatorname{CALL}
\mid d^\star=\operatorname{ANSWER}
\right),
\end{equation}

\begin{equation}
\mathrm{BTA}
=
\tfrac{1}{2}\mathrm{TCR}
+
\tfrac{1}{2}(1-\mathrm{OCR}).
\label{eq:bta}
\end{equation}
Thus TCR requires a complete correct call, OCR measures over-calling on
\answer{} examples, and BTA balances correct calling and non-calling. We also
report Decision Accuracy, Tool Accuracy on gold \call{} examples, and Argument
Exact Match conditional on correct tool selection; under this decomposition,
$\mathrm{TCR}=\mathrm{ToolAcc}\times\mathrm{ArgExact}$.

\subsection{Oracle-Guided On-Policy Distillation}

\subsubsection{Design}

\ours{} combines three components motivated by the benchmark diagnosis.

\paragraph{Reliable-State Conditioning.}
The teacher conditions on the synchronized Oracle State, representing the
policy induced by authoritative task evidence rather than by competing
historical traces.

\paragraph{Distributional Policy Transfer.}
Token-level soft targets preserve the teacher's relative preferences over
answering, calling, tool selection, and argument generation, which are
discarded by a single hard target sequence.

\paragraph{Student-State Coverage.}
A fixed gold or teacher-generated prefix may not cover the states reached by a
student conditioned on misleading history. \ours{} therefore evaluates the
teacher on prefixes generated by the current student, following on-policy
language-model and context-distillation formulations
\cite{ICLR2024_5be69a58,ye2026onpolicycontextdistillationlanguage}.

\subsubsection{Objective}

Let $\pi_\theta$ denote the student and $\pi_T$ the frozen teacher. Given paired
Polluted and Oracle inputs $(x_p,x_o)$, the student first generates
\begin{equation}
y \sim \pi_\theta(\cdot \mid x_p).
\end{equation}
For each student-generated prefix $y_{<t}$, we define
\begin{align}
q_t^{S} &= \pi_\theta(\cdot \mid x_p,y_{<t}),\\
q_t^{T} &= \pi_T(\cdot \mid x_o,y_{<t}).
\end{align}
The training objective is
\begin{equation}
\mathcal{L}_{\mathrm{OPD}}
=
\mathbb{E}_{(x_p,x_o),\,y}
\left[
\frac{1}{|y|}
\sum_{t=1}^{|y|}
D_{\mathrm{KL}}
\left(
q_t^{S}\,\|\,q_t^{T}
\right)
\right].
\label{eq:opd}
\end{equation}

We compute reverse KL over the union of the teacher and student top-16 token
supports, following reverse-KL language-model distillation
\cite{gu2024minillm}. We use temperature 1 and four student rollouts per
prompt, without an additional gold-CE objective. At deployment, only the
trained student and ordinary interaction history are required; the teacher and
Oracle State are removed.
\section{Experiments}

Table~\ref{tab:main} summarizes the central post-training comparison before
we detail the experimental protocol.

\subsection{Setup}

We train Qwen3-1.7B and Qwen3-8B for three epochs with full-parameter FSDP on
eight H20 GPUs. OPD uses learning rate $10^{-6}$, global batch size 32
prompts, four rollouts per prompt, and prompt/response limits of
12,288/1,024 tokens. SFT uses cosine learning rate $10^{-5}$ and global batch
size 256. Evaluation uses greedy decoding. Checkpoints are selected by
validation BTA before test evaluation. Detailed computational-cost measurements
are reported in Appendix~C.

Main comparisons include the untuned model, Gold-SFT, call-rebalanced and
sequence-normalized SFT, SFT\allowbreak$\rightarrow$\allowbreak DPO
\cite{rafailov2023direct}, Oracle-SeqKD, off-policy OD, and \ours.
Additional analyses vary teacher view, prefix source, teacher capacity, and
student scale. For cross-generator evaluation, we keep checkpoint selection fixed and perform
inference only on evaluation contexts independently regenerated by
Claude-Opus-4.7; no Claude-generated example is used for training, adaptation,
or model selection. Complete per-view metrics are reported in Appendix~D,
while construction-time validation and filtering details are provided in
Appendix~A.

\subsection{Diagnosing History-Induced Policy Hijacking}

Table~\ref{tab:crossmodel} shows a broad reliability gap across model scales.
Replacing Polluted history with Original or Oracle produces substantial gains,
and the Oracle advantage remains visible at 32B. The benchmark therefore
captures a persistent input-reliability bottleneck rather than a peculiarity of
one small checkpoint.

\begin{table}[t]
\centering
\small
\setlength{\tabcolsep}{4pt}
\begin{tabular}{@{}lccc@{}}
\toprule
Model & P & O & R\\
\midrule
\multicolumn{4}{c}{\textit{Test-Indist BTA $\uparrow$}}\\
\midrule
Q3-1.7B   & 0.4720 & 0.6713 & 0.7450\\
Q3-8B     & \best{0.6888} & \best{0.8416} & \best{0.8910}\\
Q3-32B    & 0.6587 & 0.7170 & 0.8773\\
DS-V4-F   & 0.5691 & 0.5829 & 0.6613\\
\midrule
\multicolumn{4}{c}{\textit{OOD-600 TCR $\uparrow$}}\\
\midrule
Q3-1.7B   & 0.6233 & 0.7367 & 0.7067\\
Q3-8B     & 0.8167 & 0.8250 & 0.8433\\
Q3-32B    & \best{0.8733} & \best{0.9183} & \best{0.9433}\\
DS-V4-F   & 0.3950 & 0.3883 & 0.4817\\
\bottomrule
\end{tabular}
\caption{Three-view evaluation across model scales. P, O, and R denote the Polluted, Original, and Oracle State views, respectively; Q3 and DS-V4-F denote Qwen3 and DeepSeek-V4-Flash. Test-Indist is evaluated with BTA, while the \call-only OOD-600 split is evaluated with TCR.}
\label{tab:crossmodel}
\end{table}

\begin{figure}[t]
\centering
\IfFileExists{fig2_causal_evidence.pdf}{
  \includegraphics[
    width=0.97\columnwidth
  ]{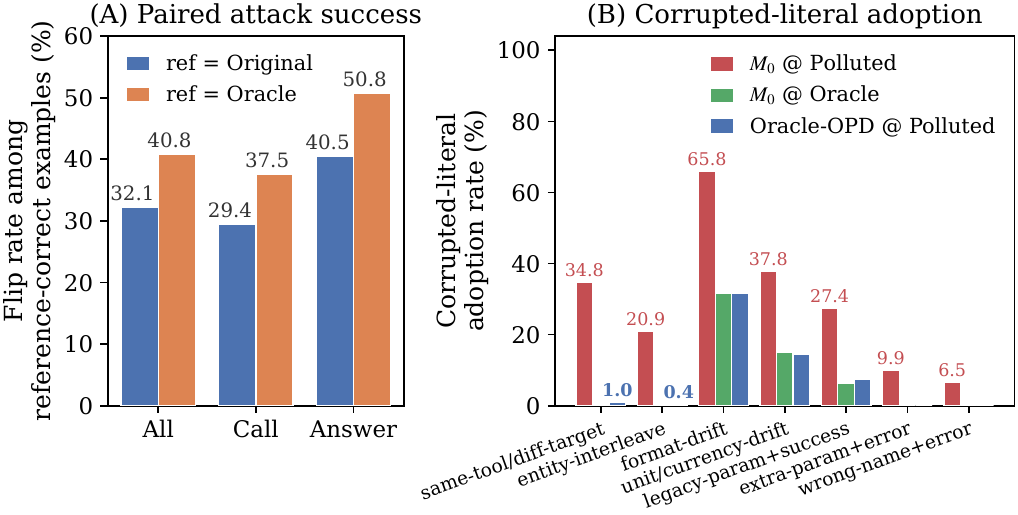}
}{
  \fbox{\parbox[c][6.0cm][c]{0.90\columnwidth}{\centering
  \textbf{Missing file: fig2\_causal\_evidence.pdf}}}
}
\caption{Paired prediction flips and corrupted-literal adoption.
(A) Flip rates from correct predictions under Original or Oracle State to
incorrect predictions under the paired Polluted view, reported for all,
gold-\call{}, and gold-\answer{} examples.
(B) Corrupted-literal adoption rates for the seven \call-side operators
with identifiable injected literals, reported for $M_0$ under Polluted and
Oracle State inputs and for \ours{} under Polluted input.}
\label{fig:causal}
\end{figure}

Aggregate gaps alone could reflect generic difficulty. We therefore condition
on examples the same model solves under a reliable reference.
Figure~\ref{fig:causal}A shows that $32.1\%$ of Qwen3-1.7B predictions correct
on Original become wrong on Polluted. The rate is $29.4\%$ for \call{} and
$40.5\%$ for \answer{}. Relative to Oracle-correct predictions, the
corresponding rates are $40.8\%$, $37.5\%$, and $50.8\%$. Pollution thus
changes behavior on tasks whose underlying policy is already available.

Figure~\ref{fig:causal}B identifies the shortcut through corrupted-literal
adoption. This analysis covers the seven \call-side operators, among the
eleven total, that inject a concrete entity, argument name, or surface form;
\answer-side operators and missing-required errors have no literal to adopt.
Under Polluted input, the base model adopts the corrupted literal in $34.8\%$
of same-tool/different-target examples, $20.9\%$ of entity-interleaving
examples, and $65.8\%$ of format-drift examples. The $M_0$-Oracle result is a
clean-state baseline rather than an Oracle error: it measures how often the
same literal would arise naturally without pollution, which can be nonzero for
plausible formats, units, or legacy aliases. After \ours{} training, adoption
falls to $1.0\%$ and $0.4\%$ on the two entity-binding operators even though
the misleading trace remains visible. The benchmark therefore exposes a
competing historical state representation, and the method teaches the student
which state should govern the current action.

\subsection{Main Results}

\begin{table*}[t]
\centering
\small
\setlength{\tabcolsep}{3pt}
\begin{tabular}{@{}lrrrrrrr@{}}
\toprule
& \multicolumn{6}{c}{Test-Indist}
& \multicolumn{1}{c}{OOD-600}\\
Method
& BTA $\uparrow$
& TCR $\uparrow$
& $1-\mathrm{OCR}$ $\uparrow$
& \shortstack{Decision Acc.\\$\uparrow$}
& \shortstack{Tool Acc.\\$\uparrow$}
& \shortstack{ArgExact$\,|\,Tool$\\$\uparrow$}
& TCR $\uparrow$\\
\midrule
$M_0$ Polluted
& 0.4720 & 0.5807 & 0.3633 & 0.7938 & 0.8993 & 0.6457 & 0.6233\\
\midrule
Gold-SFT
& 0.6628 & 0.3422 & 0.9834 & 0.6838 & 0.5370 & 0.6373 & 0.2467\\
Sampled-SFT (2:1)
& 0.6620 & 0.3392 & 0.9848 & 0.6728 & 0.5237 & 0.6477 & 0.2400\\
SeqNorm-SFT
& 0.5983 & 0.1966 & 1.0000 & 0.4760 & 0.2458 & 0.8000 & 0.1000\\
SFT$\rightarrow$DPO
& 0.6350 & 0.2700 & 1.0000 & 0.5384 & 0.3343 & 0.8076 & 0.1300\\
\midrule
Oracle-SeqKD
& 0.8229 & 0.7688 & 0.8771 & 0.9410 & 0.9066 & 0.8481 & 0.7133\\
Off-policy OD
& 0.8499 & 0.9041 & 0.7956 & 0.9376 & 0.9642 & \best{0.9377} & \best{0.7433}\\
\best{\ours}
& \best{0.8703} & \best{0.9132} & 0.8273
& \best{0.9473} & \best{0.9824} & 0.9296 & \best{0.7433}\\
\bottomrule
\end{tabular}
\caption{Main Qwen3-1.7B post-training results on \bench.
\ours{} achieves the strongest balanced tool-use policy, combining high
complete-call recall with reliable non-call restraint and obtaining the best
decision and tool accuracy. OOD-600 contains only \call-required examples
whose gold actions invoke six tools unseen during training and is therefore
evaluated with TCR. }
\label{tab:main}
\end{table*}

\paragraph{Reliable-state distillation learns the strongest balanced policy.}
Table~\ref{tab:main} shows that \ours{} reaches $0.8703$ BTA, combining
$0.9132$ complete-call recall with $0.8273$ non-call recall and obtaining
the highest decision and tool accuracy among same-size methods. Reliable-state
supervision produces a consistent progression: Oracle-SeqKD, off-policy OD,
and \ours{} reach $0.8229$, $0.8499$, and $0.8703$ BTA, respectively. Under
paired bootstrap with 10,000 resamples, \ours{} improves over off-policy OD by
$+0.0204$ BTA ($95\%$ CI $[+0.0118,+0.0295]$, $p<10^{-4}$).

Gold-SFT and its rebalanced, sequence-normalized, and DPO variants instead
concentrate in a high-restraint regime, with near-perfect non-calling but much
lower complete-call recall. The contrast shows that the balanced gain does not
follow from class reweighting, sequence normalization, or standard preference
optimization alone; the factorized method components are examined in
Table~\ref{tab:design}.

\subsection{Learning a Balanced Tool-Use Policy}

\begin{figure}[t]
\centering
\IfFileExists{fig3_strategy_plane.pdf}{
  \includegraphics[
    width=0.82\columnwidth,
    height=0.29\textheight,
    keepaspectratio
  ]{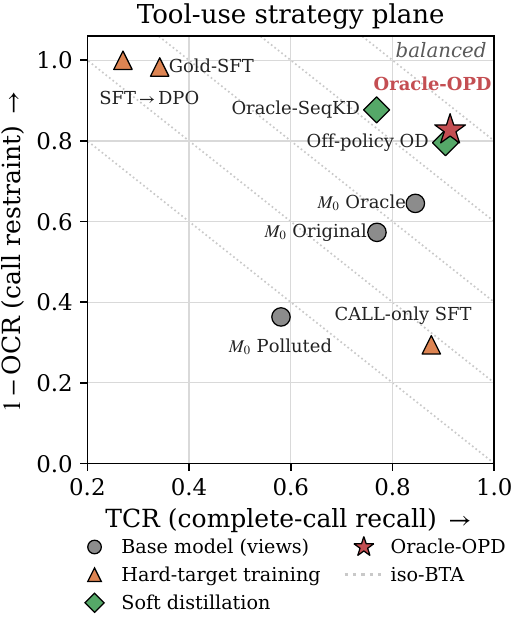}
}{
  \fbox{\parbox[c][4.4cm][c]{0.78\columnwidth}{\centering
  \textbf{Missing file: fig3\_strategy\_plane.pdf}}}
}
\caption{Tool-use strategy plane. The horizontal axis reports complete-call recall, and the vertical axis reports non-call recall, $1-\mathrm{OCR}$. Each point represents a training configuration. Dotted anti-diagonals connect points with equal BTA.}
\label{fig:strategy}
\end{figure}

Figure~\ref{fig:strategy} reveals three distinct policy regimes. Training only
on \call{} examples preserves high execution recall but provides no supervision
for when tool use should stop: CALL-only Gold-SFT reaches $0.8762$ TCR but only
$0.2942$ non-call recall. Mixed Gold-SFT learns the opposite boundary,
reaching $0.9834$ non-call recall but $0.3422$ TCR. In contrast, mixed
\ours{} reaches the upper-right region with $0.9132$ TCR and $0.8273$ non-call
recall.

This comparison clarifies both the benchmark and the method. \bench{} evaluates
call execution and call control jointly, while Oracle-conditioned soft
supervision transfers both complete-call execution and selective tool-use
control.

\subsection{Robustness Transfers to Clean Histories}

A model trained under Polluted input should learn a reliable tool-use policy
rather than a correction rule tied to the injected traces. We therefore freeze
the exact checkpoints selected for Table~\ref{tab:main} and reevaluate them
without additional training on the synchronized Original view.
Table~\ref{tab:clean_transfer} reports this cross-view evaluation, including
the same 600 held-out-function examples used in the main results.

\begin{table}[t]
\centering
\small
\setlength{\tabcolsep}{3.5pt}
\begin{tabular}{@{}lrrrr@{}}
\toprule
& \multicolumn{2}{c}{ID BTA $\uparrow$}
& \multicolumn{2}{c}{OOD TCR $\uparrow$}\\
Method & P & O & P & O\\
\midrule
\multicolumn{5}{l}{\textit{1.7B student}}\\
Gold-SFT
& 0.6628 & 0.5880 & 0.2467 & 0.0750\\
Oracle-SeqKD
& 0.8229 & 0.8082 & 0.7133 & 0.6667\\
Off-OD
& 0.8499 & 0.8205 & \best{0.7433} & 0.7333\\
\best{\ours}
& \best{0.8703} & \best{0.8450}
& \best{0.7433} & \best{0.7450}\\
\midrule
\multicolumn{5}{l}{\textit{8B student}}\\
Gold-SFT
& 0.9221 & 0.9086 & 0.7967 & 0.7650\\
\ours{} (base T)
& 0.9077 & 0.9092 & 0.8200 & \best{0.8333}\\
\best{\ours{} (O-SFT T)}
& \best{0.9301} & \best{0.9260}
& \best{0.8217} & 0.8050\\
\bottomrule
\end{tabular}
\caption{Polluted- and Original-view results of post-trained
checkpoints on Test-Indist and OOD-600.
P and O denote Polluted and Original inputs, respectively.}
\label{tab:clean_transfer}
\end{table}

Reliable-state distillation transfers to ordinary histories rather than
learning an error-specific correction. For the 1.7B student, \ours{} reaches
$0.8450$ BTA on Original input, while held-out-function TCR remains essentially
unchanged between Polluted and Original views ($0.7433$ versus $0.7450$). The
same stability holds at 8B: the base-teacher model scores $0.9077/0.9092$ BTA
on Polluted/Original inputs, and the Oracle-specialized variant scores
$0.9301/0.9260$. These results show that reliable-state supervision changes
which evidence governs the policy across both polluted and clean histories.

\subsection{Robustness across Interference Operators}

The aggregate gain extends across the mechanisms represented in the benchmark.
Table~\ref{tab:operators} reports complete-call recall over all eight
\call-side operators. The remaining three operators intervene on
\answer-required decisions and are captured by OCR and BTA in
Table~\ref{tab:main}.

\ours{} is best or tied on seven of the eight \call-side operators and obtains
the strongest operator-macro TCR. Its largest gains occur on unit or scale
drift and on the two entity-binding interventions, where the injected trace
offers a highly plausible but incorrect target for the current call. These
results show that the aggregate improvement is not driven by one template or
one interface pattern; it extends from entity selection to schema use and
argument realization.

\begin{table}[t]
\centering
\small
\setlength{\tabcolsep}{2.5pt}
\begin{tabular}{@{}lrrrr@{}}
\toprule
Operator
& $M_0$
& Gold-SFT
& Off-OD
& \ours\\
\midrule
Same tool, other target
& 50.7 & 32.6 & 90.3 & \best{91.6}\\
Entity interleaving
& 53.4 & 28.5 & 89.2 & \best{91.6}\\
Wrong name + error
& 73.5 & 42.8 & \best{91.6} & 89.3\\
Missing required
& 62.9 & 41.2 & \best{91.2} & \best{91.2}\\
Extra parameter
& 68.9 & 34.2 & 91.4 & \best{92.3}\\
\legacyop
& 60.7 & 32.9 & 90.9 & \best{91.8}\\
Format drift
& 40.0 & 35.8 & \best{93.3} & \best{93.3}\\
Unit / scale drift
& 44.5 & 24.4 & 84.0 & \best{89.1}\\
\midrule
Operator macro
& 56.8 & 34.1 & 90.3 & \best{91.3}\\
\bottomrule
\end{tabular}
\caption{Complete-call recall across the eight \call-side interference operators. Values report TCR (\%) for each operator. The final row reports the operator-macro result.}
\label{tab:operators}
\end{table}

\paragraph{Cross-generator transfer.}
To test generator dependence, Claude-Opus-4.7 independently regenerates
evaluation pollution traces and Oracle States, while all checkpoints remain
trained only on DeepSeek-V4-Flash data. Without retraining, \ours{} remains the
strongest 1.7B post-trained model under Polluted input ($0.814$ BTA versus
$0.792$ for off-policy OD and $0.761$ for Oracle-SeqKD). The
Oracle-specialized 8B variant likewise reaches $0.924$ BTA. Thus both the
three-view diagnosis and the benefit of reliable-state distillation transfer
across generators. Full results are provided in Appendix~E.

\subsection{What Drives Reliable-State Policy Transfer?}

The method combines three choices: a reliable teacher view, distributional
rather than sequence-only supervision, and direct coverage of student-visited
prefixes. Table~\ref{tab:design} factorizes these choices.

\begin{table}[t]
\centering
\small
\setlength{\tabcolsep}{1.5pt}
\begin{tabular}{@{}lcccr@{}}
\toprule
Method
& \shortstack{Teacher\\view}
& Target
& Prefix
& BTA $\uparrow$\\
\midrule
Original-OPD
& Original & Soft & Student & 0.7562\\
Oracle-SeqKD
& Oracle & Hard & Teacher & 0.8229\\
Off-policy OD
& Oracle & Soft & Gold & 0.8499\\
Teacher-prefix OD
& Oracle & Soft & Teacher & 0.8479\\
Curriculum OPD
& Oracle & Soft & T$\rightarrow$S & 0.7923\\
\best{\ours}
& Oracle & Soft & Student & \best{0.8703}\\
\bottomrule
\end{tabular}
\caption{Factorized design comparison on Test-Indist. The configurations vary
the teacher input view, supervision target, and prefix source.
T$\rightarrow$S denotes a teacher-to-student prefix curriculum.
The final column reports BTA.}
\label{tab:design}
\end{table}

\paragraph{Reliable conditioning determines which state governs the policy.}
With the target distribution and student-prefix training fixed, replacing the
Original teacher view with Oracle State raises BTA from $0.7562$ to $0.8703$.
The gain shows that a decision-focused reliable state representation provides
stronger supervision than an unstructured clean trajectory when the student
must act under competing historical evidence.

\paragraph{Soft supervision transfers more than one decoded action.}
Under the Oracle teacher view, token-level off-policy distillation reaches
$0.8499$ BTA, compared with $0.8229$ for Oracle-SeqKD. The teacher
distribution therefore conveys useful preferences over whether to answer or
call, which tool to select, and how to realize the arguments beyond the
information retained by one hard sequence.

\paragraph{Direct student-state coverage completes the transfer.}
Using student-generated prefixes raises BTA from $0.8479$ for teacher-prefix
OD and $0.8499$ for gold-prefix off-policy OD to $0.8703$. A gradual
teacher-to-student curriculum reaches $0.7923$, while direct student-prefix
coverage throughout training produces the strongest balanced policy. Together,
these comparisons support the full design: Oracle State supplies the
authoritative policy, soft targets preserve its structure, and on-policy
prefixes align supervision with deployment-time behavior.

\subsection{Scaling Teacher and Student Capacity}

Table~\ref{tab:scale} evaluates two complementary scaling axes. The upper block
holds the deployed student fixed at 1.7B and changes only teacher capacity.
Replacing the 1.7B teacher with an 8B teacher raises BTA from $0.8703$ to
$0.9193$, while retaining the same compact 1.7B deployment model. Reliable
Oracle-view policy knowledge can therefore be transferred across model sizes
without increasing inference cost.

The lower block scales the student itself. At 8B, Gold-SFT provides a strong
reference policy at $0.9221$ BTA. An Oracle-specialized teacher further raises
\ours{} to $0.9301$ BTA and $0.8217$ OOD TCR. Reliable-state distillation
therefore benefits from both stronger teachers and more capable students, while
the cross-size configuration transfers stronger policy quality into a smaller
deployment model.

\begin{table}[t]
\centering
\small
\setlength{\tabcolsep}{2.5pt}
\begin{tabular}{@{}c>{\raggedright\arraybackslash}p{0.40\columnwidth}rr@{}}
\toprule
Student & Training / teacher & BTA $\uparrow$ & OOD TCR $\uparrow$\\
\midrule
\multirow{3}{*}{1.7B}
& Gold-SFT & 0.6628 & 0.2467\\
& \ours, 1.7B teacher & 0.8703 & 0.7433\\
& \ours, 8B teacher & \best{0.9193} & \best{0.7500}\\
\midrule
\multirow{3}{*}{8B}
& Gold-SFT & 0.9221 & 0.7967\\
& \ours, 8B base teacher & 0.9077 & 0.8200\\
& \ours, Oracle-SFT teacher & \best{0.9301} & \best{0.8217}\\
\bottomrule
\end{tabular}
\caption{Teacher- and student-scale comparison. The upper block fixes the deployed student at 1.7B and varies the training method and teacher size. The lower block evaluates an 8B student with Gold-SFT, a base-model teacher, and an Oracle-SFT teacher. Results are reported with Test-Indist BTA and OOD-600 TCR.}
\label{tab:scale}
\end{table}

\subsection{External Capability, Call-Control, and Noise-Robustness Transfer}

Table~\ref{tab:external} evaluates the 1.7B deployment models on BFCL v4
Non-Live and When2Call \cite{patil2025the,ross-etal-2025-when2call}. BFCL measures broad
function-calling capability across irrelevance, multiple and parallel calls,
and three programming-language subsets. When2Call evaluates whether a model
should invoke a tool, request information, or recognize that the available
tools are insufficient. Its test split contains no gold \textsc{Direct}
examples, so Macro-F1 averages the three supported classes.

\begin{table}[t]
\centering
\small
\setlength{\tabcolsep}{4pt}
\begin{tabular}{@{}lrrr@{}}
\toprule
Method
& BFCL $\uparrow$
& W2C $\uparrow$
& Halluc. $\downarrow$\\
\midrule
$M_0$              & 79.58 & 0.2701 & 0.5000\\
Gold-SFT           & 55.55 & 0.2995 & 0.4767\\
Off-policy OD      & 79.19 & 0.2842 & 0.4535\\
\ours              & 79.77 & 0.3003 & 0.3915\\
\ours, 8B teacher  & \best{80.45} & \best{0.3294} & \best{0.3411}\\
\bottomrule
\end{tabular}
\caption{External evaluation of the 1.7B deployment models. BFCL and W2C denote their respective macro scores. Hallucination is the rate of predicting \textsc{Tool Call} on \textsc{Cannot Answer} examples.}
\label{tab:external}
\end{table}

\paragraph{Broad function-calling capability is preserved.}
The 8B-teacher 1.7B student reaches $80.45$ BFCL macro accuracy, compared with
$79.58$ for the untuned model, while \ours{} with the same-size teacher reaches
$79.77$. Reliable-state specialization therefore preserves broad
function-calling capability rather than restricting gains to \bench{}.

\paragraph{Call-control behavior transfers beyond the benchmark.}
The cross-size student obtains the best F1 score in all three supported
When2Call classes and raises Macro-F1 from $0.2701$ to $0.3294$. It also
reduces unsupported tool invocation from $0.5000$ to $0.3411$. These gains
occur on a benchmark without the controlled history interventions used in
training, providing independent evidence that reliable-state distillation
improves the underlying tool-use policy.

\paragraph{Noise robustness transfers beyond tool-use histories.}
We further evaluate the fixed checkpoints zero-shot on the 7,405-example
HotpotQA-Distractor validation set \cite{yang-etal-2018-hotpotqa}, comparing answer generation from the full
ten-paragraph distractor context with generation from supporting paragraphs
only. The 1.7B student distilled from an 8B teacher reaches $0.5015$ F1 under
noisy context, exceeding the next-best trained checkpoint by $2.97$ F1 points,
and obtains the smallest clean-to-noisy F1 gap among the trained 1.7B models
($0.1596$). This result extends the benefit of reliable-state distillation to
irrelevant-context robustness outside tool-use interactions. Full EM, F1,
model-scale, and significance results are reported in Appendix~F.

\paragraph{Prompt-only control.}
A direct instruction to ignore failed or irrelevant history raises BTA only
from $0.4720$ to $0.4849$, compared with $0.8703$ for learned reliable-state
policy transfer. The benchmark therefore requires the model to discriminate
which historical state remains authoritative rather than merely acknowledge a
general warning about unreliable context.
\section{Limitations}

\bench{} focuses on controlled next-action interventions over airline and
retail trajectories. Extending the framework to naturally occurring production
histories and end-to-end interaction would further broaden its coverage.
\ours{} additionally uses Oracle State and on-policy rollouts during training,
while deployment requires only the distilled student and ordinary interaction
history.

\section{Conclusion}

Robust tool use requires models to identify which historical state remains
authoritative for the current request. \bench{} makes this capability
measurable through synchronized, gold-preserving Original, Polluted, and Oracle
views, revealing that misleading traces can redirect policies that models
otherwise execute correctly. \ours{} transfers an Oracle-conditioned policy to
polluted-context student rollouts through soft supervision on student-visited
prefixes. It yields the strongest balanced policy on Qwen3-1.7B, transfers across clean
histories and independently generated evaluation contexts, generalizes across
interference operators and unseen functions, and preserves noise robustness
on zero-shot multi-hop QA. Its benefits improve further with stronger teachers. These results establish history reliability as a distinct
and tractable training target for multi-turn tool-using agents.



\bibliography{references}

\clearpage
\appendix
\setlength{\emergencystretch}{1.5em}

\section{Benchmark Construction and Validation}
\label{app:construction}

This appendix specifies the construction procedure for \bench{} and maps its
components to the submitted code and data supplement. Starting from APIGen-MT
source trajectories, the pipeline constructs synchronized Original, Polluted,
and Oracle State views while preserving the current tools, latest request, and
gold next action. Source trajectories supply the gold calls and answers; model
generation is restricted to the historical intervention and reliable-state
summary. Structural, operator-specific, decision-preservation, and Oracle
checks are applied before inclusion.

We use the airline and retail domains because they combine persistent
multi-turn state, recurring entity identifiers, heterogeneous tool schemas,
and substantive \call{} and \answer{} decisions within a common trajectory
format. This combination supports controlled history interventions while
keeping the current request, available tools, and gold next action fixed.

\subsection{Construction pipeline}

The submitted supplement provides the executable construction flow
through five standalone scripts and one end-to-end driver. Table~\ref{tab:app-pipeline}
summarizes the responsibility and principal output of each stage.

\begin{table*}[t]
\centering
\small
\setlength{\tabcolsep}{5pt}
\renewcommand{\arraystretch}{1.08}
\begin{tabularx}{\textwidth}{@{}c>{\raggedright\arraybackslash}p{0.20\textwidth}Y>{\raggedright\arraybackslash}p{0.17\textwidth}@{}}
\toprule
Stage & Entry point & Responsibility & Output\\
\midrule
1 & \codeid{1\_split\_seeds.py}
  & trajectory-level split assignment and held-out-tool routing
  & \codeid{seed\_split.json}\\
2 & \codeid{2\_build\_pollution.py}
  & \call{} and \answer{} decision extraction and Polluted-view generation
  & \codeid{pollution.jsonl}\\
3 & \codeid{3\_validate.py}
  & standalone structural re-validation of every generated record
  & \codeid{validated.jsonl}\\
4 & \codeid{4\_build\_oracle.py}
  & Oracle State generation and recoverability checks
  & \codeid{with\_oracle.jsonl}\\
5 & \codeid{5\_make\_splits.py}
  & final routing into train, validation, Test-Indist, and OOD-600
  & four split files\\
\bottomrule
\end{tabularx}
\caption{Benchmark-construction pipeline included in the submitted supplement.}
\label{tab:app-pipeline}
\end{table*}

The complete flow is invoked by \codeid{run\_pipeline.sh}. The third-party
\codeid{apigen-mt\_5k.json} source file is obtained from the APIGen-MT release
and is not redistributed. The pipeline produces \codeid{train.jsonl},
\codeid{val.jsonl}, \codeid{test\_indist.jsonl}, and
\codeid{test\_ood\_600.jsonl}; exact invocation and generator configuration are
recorded in the submitted README and source files.

\subsection{Source normalization and\\decision extraction}

Each APIGen-MT trajectory first passes a deterministic source sanity check
covering the system policy, tool inventory, conversation structure, and parsed
function calls. Native messages are then converted to a common OpenAI-format
representation. Throughout this appendix, Original denotes the resulting
intervention-free source prefix. The same normalization is shared by all
synchronized views, and the intervention pipeline does not edit, delete, or
reorder the normalized source messages.

For a \call-required item, the construction code selects the trajectory's last
non-\texttt{think} tool call when that call is immediately preceded by a user
turn. The preceding user message is the latest request $u_t$, and the source
function name and fully parsed argument object define the gold action
$(f^\star,\mathbf z^\star)$. The intervention region ends strictly before
$u_t$, so the latest request remains the final message in both Original and
Polluted inputs.

For an \answer-required item, the code identifies substantive user turns
followed by a plain assistant response with no tool call. A fixed classification
prompt labels the source response as \textsc{Clarify},
\textsc{Answer-from-Evidence}, \textsc{Unsupported}, or \textsc{Social}; the
first three decision types are retained. The classifier also records the
missing information, supporting evidence, or unsupported reason needed for
subsequent validation. At most one decision point per retained subtype is kept
from each source trajectory.

\subsection{Trajectory-level splits and held-out functions}

Split assignment is performed at the source-trajectory level before pollution
generation, and every later record inherits that assignment. We first route a
trajectory to OOD when its terminal gold \call{} uses one of the six held-out
functions listed below. The remaining trajectories are partitioned jointly into
train, validation, and Test-Indist by deterministic greedy target-proportion
packing with $70\%/10\%/20\%$ trajectory-level targets. Each source trajectory
is treated as one indivisible group. Following a fixed trajectory ordering, the
procedure assigns each group to the split that best preserves the target
proportions, with deterministic tie-breaking. Consequently, sibling decision
points, recurring entities, and nearby prefixes cannot cross split boundaries.
A trajectory enters the OOD bucket when its terminal gold \call{} invokes one
of the following six held-out functions:
\begin{quote}\small\raggedright
\codeid{update\_reservation\_passengers},\\
\codeid{get\_reservation\_details},\\
\codeid{return\_delivered\_order\_items},\\
\codeid{modify\_pending\_order\_address},\\
\codeid{modify\_pending\_order\_payment}, and\\
\codeid{get\_product\_details}.
\end{quote}
These functions have related same-family interfaces in the non-OOD data, so the
split tests transfer to unseen gold functions rather than an unrelated tool
family. OOD trajectories contribute only their \call-required decision point;
non-call examples from those trajectories are not emitted. The final routing
stage explicitly checks that no trajectory identifier appears in two splits and
that no OOD gold function appears as a gold target in train, validation, or
Test-Indist.

\subsection{Polluted-view generation for \call{} decisions}

The \call-side constructor receives the normalized source prefix, current tool
schemas, source gold continuation, assigned operator, and legal insertion
indices. Conditioning on the source continuation keeps the inserted precedent
consistent with the synchronized gold decision; the continuation is never
inserted into an evaluation input. The constructor follows the output contract
below:
\begin{enumerate}
    \item return one strict JSON object containing \codeid{insert\_idx},
    \texttt{injection}, and operator-specific \texttt{literals};
    \item serialize tool arguments and tool results as JSON strings in valid
    OpenAI-format messages;
    \item insert only at a legal model-turn boundary and never split an
    assistant-call/tool-result pair;
    \item preserve every source message and keep $u_t$ final;
    \item avoid meta-language that would reveal that the trace is synthetic or
    intentionally misleading;
    \item instantiate every required literal and use neutral
    \codeid{call\_<12 hex>} identifiers; and
    \item provide no corrective hint that teaches the model the gold interface
    or gold entity for free.
\end{enumerate}

The returned object follows the fixed schema below:
\begin{CodeBlock}
{
  "insert_idx": <integer>,
  "injection": [<OpenAI-format messages>],
  "literals": {<operator-specific fields>},
  "rationale": "<optional; not used by validation>"
}
\end{CodeBlock}

One accepted \call-side variant is retained per source trajectory. Generation
is followed by operator-specific validation; invalid or inapplicable outputs
are discarded and construction proceeds with another applicable intervention.
The exact operator scheduling and retry behavior is included in the submitted
pipeline.

\subsection{Complete \call-side contracts}

Table~\ref{tab:app-call-contracts} records the generated trace and the decisive
acceptance conditions for each \call-side operator. These conditions are
implemented as operator-specific deterministic validators after the common
message-format checks.

\begin{table*}[t]
\centering
\footnotesize
\setlength{\tabcolsep}{5pt}
\renewcommand{\arraystretch}{1.08}
\begin{tabularx}{\textwidth}{@{}>{\raggedright\arraybackslash}p{0.16\textwidth}YY@{}}
\toprule
Operator & Generated precedent & Acceptance conditions\\
\midrule
Same tool, other target
& A call to the gold function with one entity identifier replaced by a
same-format alternative, followed by a result rewritten for that entity.
& The function equals the gold function; the fake and real identifiers differ;
the fake identifier appears in arguments and result; the real identifier does
not appear in the injected result.\\
Entity interleaving
& Two or three call--result pairs to the gold function, each using a different
same-format distractor entity.
& The declared pair count matches the message count; all distractor identifiers
are distinct from the gold entity; every result is bound to its paired
distractor and excludes the real identifier.\\
Wrong name + error
& A two-message call--result pair in which one real parameter is replaced by a
non-schema name and the API rejects that name.
& The wrong name is present in the call and error, is absent from the tool
schema, and the error contains neither a real parameter name nor a corrective
phrase such as ``did you mean'' or ``valid parameters.''\\
Legacy parameter + success
& The same invalid-name call followed by a success-looking result carrying an
opaque, natural-looking tracking value.
& The invalid name is absent from the schema; the result contains the declared
tracking value and contains no error or synthetic-data marker.\\
Missing required
& A call omitting exactly one required field, followed by a generic missing-field
rejection.
& The omitted field is required by the schema, is absent from the call, and is
not named by the tool result; the remaining required arguments are retained.\\
Extra parameter
& A valid call augmented with one unsupported field, followed by an error naming
only that field.
& The extra key is absent from the schema, present in the call and rejection,
and the rejection does not enumerate accepted fields.\\
Format drift
& A call to the gold function in which one argument uses a plausible alternative
surface form, followed by a success-looking result.
& Drifted and canonical values differ; the declared nested parameter path resolves
to the drifted value; the result excludes the canonical value and any
normalization hint.\\
Unit / scale drift
& A call to the gold function with one numeric, currency, unit, magnitude, or
timezone interpretation changed, followed by a success-looking result.
& Drifted and canonical values differ at the declared argument path; the result
contains neither the canonical interpretation nor conversion commentary.\\
\bottomrule
\end{tabularx}
\caption{Complete semantic contracts for the eight \call-side operators. Common JSON, call-id, insertion-boundary, and turn-alternation checks apply to every row.}
\label{tab:app-call-contracts}
\end{table*}

\subsection{Polluted-view generation for \answer{} decisions}

The \answer-side pipeline separates source labeling from intervention
generation. A subtype-specific trace is inserted immediately before the
unchanged latest request:
\begin{itemize}
    \item \textsc{Clarify} receives a prior reassurance or decision-irrelevant
    read that does not supply the missing identity or required slot.
    \item \textsc{Answer-from-Evidence} receives a historical read associated
    with a different entity or another decision-irrelevant observation.
    \item \textsc{Unsupported} receives a plausible success-looking historical
    exchange involving a capability absent from the current tool inventory.
\end{itemize}
These realizations instantiate the three Decision-State operators used in the
paper: unsupported completion, decision-irrelevant evidence, and an unavailable
historical capability. Generated traces must satisfy the declared tool-schema,
entity-binding, and missing-information constraints before entering the
preservation gate below.

\subsection{Preservation gate for \answer{} decisions}

After an \answer-side trace passes the deterministic structural checks, an
independent zero-temperature decision auditor receives only the Polluted
conversation and the current tool inventory. It must return strict JSON with an
\texttt{action} in \{\codeid{CALL\_TOOL},\codeid{NO\_CALL}\} and a list
\codeid{now\_present} of previously missing items that the intervention has
newly supplied. A record is retained only when the auditor returns
\codeid{NO\_CALL} and \codeid{now\_present} is empty. This gate prevents a
plausible historical trace from silently completing a required slot or otherwise
changing the source non-call decision. The source gold answer itself is never
rewritten by this gate.

This design gives the two decision types parallel but appropriately specialized
validation: \call-side examples use schema-aware deterministic operator
validators tied to the fixed source call, while \answer-side examples add an
independent semantic decision gate because non-call preservation cannot be
reduced to argument-schema checks alone.

\subsection{Oracle State prompt and validation}

Oracle State is generated once for each unique
(\codeid{seed\_id}, \codeid{cut\_idx}) decision point and reused across all
Polluted variants of that point. The constructor receives the Original history,
current system policy, current tool schemas, latest request, and source gold
action. The gold action is used only during construction and is never included
in the rendered Oracle input. The output schema is:
\begin{CodeBlock}
{
  "current_user_intent": "...",
  "known_slots": {"field": "value", ...},
  "verified_observations": ["..."],
  "relevant_policy": ["..."],
  "missing_information": ["..."],
  "evidence_status": "needs_tool_call | sufficient_to_answer |
                      missing_info | unsupported",
  "irrelevant_history_to_drop": ["..."]
}
\end{CodeBlock}
The prompt requires concrete identifiers, dates, names, and amounts to be copied
verbatim, while forbidding function names, complete calls, or action directives
such as ``should call'' or ``should query.'' The rendered Oracle view preserves
the original system policy and presents the state summary together with the
unchanged latest request.

Oracle States are retained only after non-leakage, faithfulness, and
recoverability checks verify that the summary omits the gold function name,
contains the reliable information required for the decision, and introduces no
unsupported evidence. Records that do not satisfy these checks are excluded.
The exact executable prompt and validation implementation are included in the
submitted \codeid{oracle.py} source.
\subsection{Standalone structural validation}

The third pipeline stage re-validates the entire generated file independently
of the online checks used during generation. A record is retained only if:
\begin{enumerate}
    \item every message has an allowed role and every tool message and tool call
    contains the required OpenAI-format fields;
    \item the spliced conversation satisfies user/assistant turn-alternation
    rules, including call--result chaining within one assistant turn;
    \item the latest request is present verbatim and, for \call-required items,
    remains the final input message;
    \item the gold object is a well-formed complete call or a non-empty source
    answer; and
    \item the Original message sequence is an in-order subsequence of the
    Polluted sequence, establishing that no source message was edited, deleted,
    or reordered.
\end{enumerate}

The common checks are complemented by the operator-specific conditions in
Table~\ref{tab:app-call-contracts}, the \answer-side preservation gate, and the
Oracle non-leakage and recoverability checks. Only records satisfying the full
construction contract enter the final benchmark splits.

\subsection{Submitted record schema}

Every final JSONL line is self-contained. Table~\ref{tab:app-record-schema}
lists the principal fields in the submitted data artifact.

\begin{table*}[t]
\centering
\small
\setlength{\tabcolsep}{6pt}
\renewcommand{\arraystretch}{1.08}
\begin{tabularx}{\textwidth}{@{}>{\raggedright\arraybackslash}p{0.23\textwidth}Y@{}}
\toprule
Field & Content\\
\midrule
\codeid{id} & Stable example identifier containing the source seed and decision cut.\\
\codeid{original\_context} & Untouched normalized source prefix ending at $u_t$.\\
\codeid{polluted\_context} & Source prefix plus one accepted historical intervention.\\
\codeid{oracle\_state\_view} & System policy and reliable state summary paired with the unchanged request.\\
\codeid{current\_tools} & Current OpenAI-format function schemas.\\
\codeid{u\_t} & Latest user request, identical across the synchronized views.\\
\codeid{gold\_action} & Complete source call or source non-call answer.\\
\codeid{sample\_kind} & Internal decision-type field used by the data loader; maps to \call-required or \answer-required.\\
\codeid{decision\_subtype} & \textsc{Clarify}, \textsc{Answer-from-Evidence}, or \textsc{Unsupported} for non-call items.\\
\codeid{pollution\_type} & Coarse storage-level family; the realized paper-facing operator is recorded in \codeid{source\_span}.\\
\codeid{source\_span} & Realized operator, insertion position, injected literals, and generated call identifiers.\\
\codeid{provenance} & Source seed, trajectory-level split, domain, decision cut, and generator engine.\\
\codeid{oracle\_meta} & Oracle attempt count, evidence status, and structured extracted state.\\
\bottomrule
\end{tabularx}
\caption{Principal fields in each submitted synchronized benchmark record.}
\label{tab:app-record-schema}
\end{table*}

\subsection{Operator taxonomy and prompt package}

The benchmark uses eleven semantic interventions. Their grouping is fixed
across the construction code, audit rubric, and operator-level evaluation.

\begin{table*}[t]
\centering
\small
\setlength{\tabcolsep}{6pt}
\renewcommand{\arraystretch}{1.08}
\begin{tabularx}{\textwidth}{@{}>{\raggedright\arraybackslash}p{0.20\textwidth}Y@{}}
\toprule
Family & Operators\\
\midrule
Decision State
& unsupported completion claim; decision-irrelevant read; decommissioned historical tool\\
Entity Binding
& same tool with a different target; interleaved competing entities\\
Interface Execution
& wrong parameter name with rejection; legacy parameter with apparent success;
missing required argument; extra parameter; format drift; unit or scale drift\\
\bottomrule
\end{tabularx}
\caption{The eleven benchmark operators and their semantic grouping. Identity
reassurance and read-only side-query realizations are implementation alternatives
for decision-state interference rather than additional benchmark operators.}
\label{tab:app-operator-map}
\end{table*}

Each generation prompt combines a fixed shared contract, an operator-specific
clause, the serialized source prefix and current tool schemas, and a strict
output schema. The \call-side package returns an insertion index, OpenAI-format
messages, and operator-specific literals; the \answer-side package contains the
subtype classifier, intervention templates, and preservation auditor; and the
Oracle package returns the structured state object specified above.
Tables~\ref{tab:app-call-contracts} and
\ref{tab:app-prompt-clauses} specify the decision-relevant prompt contract.
The exact executable prompt strings are included in the submitted
\codeid{operators\_call.py}, \codeid{operators\_answer.py}, and
\codeid{oracle.py} files.

\subsection{Prompt contracts and executable templates}

\paragraph{Shared \call-side constructor.}
Every \call-side operator appends its semantic clause to the same constructor
contract. The compact specification below records the common serialization and
preservation requirements; the submitted source contains the executable
template.

\begin{CodeBlock}
Return one strict JSON object:
{
  "insert_idx": <integer>,
  "injection": [<OpenAI-format messages>],
  "literals": {<operator-specific fields>}
}

- Use valid assistant tool-call and paired tool-result messages.
- Insert only within the declared region; never split a call/result pair.
- Preserve every source message and leave the latest request unchanged.
- Instantiate every required operator literal in the injected trace.
- Do not include meta-language or a corrective hint revealing the
  canonical tool interface, entity, format, or value.
\end{CodeBlock}

The operator clause identifies the controlled precedent to construct.
Table~\ref{tab:app-prompt-clauses} gives the clause-level contract for all eight
\call-side operators; Table~\ref{tab:app-call-contracts} gives the corresponding
acceptance conditions.

\begin{table*}[t]
\centering
\footnotesize
\setlength{\tabcolsep}{5pt}
\renewcommand{\arraystretch}{1.06}
\begin{tabularx}{\textwidth}{@{}>{\raggedright\arraybackslash}p{0.20\textwidth}Y@{}}
\toprule
Operator & Prompt clause\\
\midrule
Same tool, other target
& Invoke the gold function on a different same-format entity and rewrite the paired result consistently for that entity, excluding the current target.\\
Entity interleaving
& Construct two or three call--result pairs to the gold function, each bound consistently to a distinct same-format distractor entity.\\
Wrong name + error
& Rename one schema parameter to an unsupported name and return a rejection that names only that form, without revealing the canonical parameter.\\
Legacy parameter + success
& Use an unsupported legacy-looking parameter name and pair it with a success-looking result carrying an opaque tracking value.\\
Missing required
& Omit exactly one required field while retaining the remaining required arguments, and return a generic rejection without naming the missing field.\\
Extra parameter
& Add one unsupported field to an otherwise valid call and return a rejection that names only the added field.\\
Format drift
& Express one argument in a plausible alternative surface form and return a success-looking result without normalization commentary.\\
Unit / scale drift
& Change one numeric, currency, unit, magnitude, or timezone interpretation and return a success-looking result without the canonical interpretation.\\
\bottomrule
\end{tabularx}
\caption{Clause-level prompt contracts for the eight \call-side operators. Each clause is paired with the deterministic acceptance conditions in Table~\ref{tab:app-call-contracts}; exact executable strings are included in the submitted code supplement.}
\label{tab:app-prompt-clauses}
\end{table*}

\paragraph{\answer-side classification and preservation.}
The \answer-side classifier labels the unchanged source reply before an
intervention is generated; a separate auditor then evaluates the completed
Polluted context without seeing the source answer. Their compact contracts are:

\begin{CodeBlock}
Classifier output:
{
  "keep": true|false,
  "decision_type": "CLARIFY" | "ANSWER_FROM_EVIDENCE" |
                   "UNSUPPORTED" | "SOCIAL",
  "missing_slots": [...],
  "answer_evidence": [...],
  "unsupported_reason": "..."
}

Independent preservation audit:
{
  "action": "CALL_TOOL" | "NO_CALL",
  "now_present": [<newly supplied previously-missing items>]
}
Retain the intervention only when action is NO_CALL and now_present is empty.
A prose reassurance does not count as a concrete identity or required slot.
\end{CodeBlock}

This separation lets the generator produce a locally natural trace while the
acceptance decision remains tied to whether the original non-call action is
preserved.

\paragraph{Oracle State extraction.}
The Oracle prompt compresses the untouched Original view into reliable state
rather than an action label:

\begin{CodeBlock}
Return strict JSON:
{
  "current_user_intent": "...",
  "known_slots": {"field": "value", ...},
  "verified_observations": [...],
  "relevant_policy": [...],
  "missing_information": [...],
  "evidence_status": "needs_tool_call" | "sufficient_to_answer" |
                     "missing_info" | "unsupported",
  "irrelevant_history_to_drop": [...]
}

- Do not name a tool or state the next action.
- Copy concrete identifiers, dates, names, and amounts verbatim.
- Include only facts supported by the Original history or system policy.
\end{CodeBlock}

The Oracle checks reject tool-name leakage and verify that the reliable
arguments or answer evidence needed for the source decision remain recoverable
from the rendered state.

\subsection{Deterministic execution and submitted artifacts}

Non-LLM routing is deterministic under the submitted configuration. Split
assignment uses trajectory-level greedy packing with deterministic tie-breaking,
and per-example construction choices are derived from stable trajectory and
example identifiers. The generator client uses a request cache keyed by the
complete generation configuration, allowing each pipeline stage to be rerun
from its serialized predecessor under the same model endpoint and request
settings.

The code supplement includes the five construction stages, shared normalization
and validation utilities, complete \call- and \answer-side prompt strings,
Oracle State construction, audit utilities, the end-to-end driver, and a README
mapping the package to the paper. This material provides the preprocessing and
benchmark-construction implementation referenced by the reproducibility
checklist.

\section{Benchmark Audit Protocols}
\label{app:audit}

We conduct two complementary audits: a focused adjudicated review of 200
examples and a stratified 1,000-example cross-audit in which human review and
Claude-Opus-4.7 apply the same criteria independently.

\subsection{Audit criteria}

The audits evaluate four primary dimensions:
\begin{enumerate}
    \item \textbf{D1: gold validity.} The supplied next action remains the
    unique correct action under both Original and Polluted input.
    \item \textbf{D2: Oracle non-leakage.} Oracle contains neither a literal
    gold-function name nor sufficient semantic scaffolding to trivially
    reconstruct a complete gold call.
    \item \textbf{D3: Oracle sufficiency and faithfulness.} Oracle contains all
    reliable information required for the decision and introduces no evidence
    absent from the source trajectory or system policy.
    \item \textbf{D4: pollution naturalness.} The inserted trace is locally
    coherent and plausible as an interaction-history artifact, rated on a
    five-point scale.
\end{enumerate}
The larger audit counts D1 as passed when the Polluted-view gold label is
correct, D2 when no literal or semantic leakage is identified, D3 only when
both sufficiency and faithfulness pass, and D4 when naturalness is at least 3.
A joint pass requires all four dimensions.

\subsection{Focused 200-example audit}

The focused sample contains 100 \call{} and 100 \answer{} examples, including
160 from Test-Indist and 40 from OOD-600. It includes interventions spanning
the benchmark's Decision-State, Entity-Binding, and Interface-Execution
categories. Two annotators label every item independently, and a third
adjudicator resolves disagreements.

\begin{table*}[t]
\centering
\small
\setlength{\tabcolsep}{8pt}
\begin{tabular}{@{}llccc@{}}
\toprule
Criterion & Subset & Ann. 1 & Ann. 2 & Adjudicated\\
\midrule
Gold correctness & All & 97.5\% & 96.5\% & 97.0\%\\
Gold correctness & \call{} & 96.0\% & 94.0\% & 95.0\%\\
Gold correctness & \answer{} & 99.0\% & 99.0\% & 99.0\%\\
Oracle sufficiency & All & 98.0\% & 99.0\% & 99.0\%\\
Oracle sufficiency & \call{} & 98.0\% & 99.0\% & 99.0\%\\
Oracle sufficiency & \answer{} & 98.0\% & 99.0\% & 99.0\%\\
Literal gold-tool leakage & 100 \call{} items & 0.0\% & 0.0\% & 0.0\%\\
Gold-label agreement & All & \multicolumn{2}{c}{98.0\% (196/200)} & --\\
\bottomrule
\end{tabular}
\caption{Focused adjudicated human audit. The sample is balanced across \call{} and \answer{} decisions.}
\label{tab:app-audit200}
\end{table*}

\subsection{Stratified 1,000-example cross-audit}

The larger audit contains 650 Test-Indist, 150 OOD-600, 100 Train, and
100 Validation examples. It is stratified across both decision types, all three
non-call subtypes, and all eleven interference operators.
\begin{table}[t]
\centering
\small
\setlength{\tabcolsep}{4pt}
\begin{tabular}{@{}lrrrrr@{}}
\toprule
Reviewer & D1 & D2 & D3 & D4 & Joint\\
\midrule
Claude-Opus-4.7 & 97.0\% & 99.6\% & 96.6\% & 96.0\% & 91.6\%\\
Human & 98.0\% & 99.6\% & 99.0\% & 96.9\% & 93.7\%\\
\bottomrule
\end{tabular}
\caption{Per-dimension and joint acceptance in the 1,000-example cross-audit. D3 requires both sufficiency and faithfulness; D4 requires a naturalness rating of at least 3.}
\label{tab:app-audit1000-pass}
\end{table}

Joint pass/fail decisions agree on 925 of 1,000 examples (92.5\%).
Dimension-level agreement is 96.6\% for D1, 100.0\% for D2, 97.0\% for D3,
and 96.5\% for D4.

\section{Training Configuration and\\Computational Cost}
\label{app:compute}

All reported checkpoints use full-parameter FSDP on eight NVIDIA H20 GPUs.
The common settings are summarized first, followed by the controlled development
choices, method-specific configurations, randomness, and execution environment.
Checkpoint selection uses validation BTA throughout; the selected checkpoint is
then reused unchanged for every reported test and transfer evaluation.

\subsection{Common training and evaluation settings}

\begin{table}[t]
\centering
\small
\setlength{\tabcolsep}{4pt}
\renewcommand{\arraystretch}{1.05}
\begin{tabularx}{\columnwidth}{@{}Ycc@{}}
\toprule
Setting & Qwen3-1.7B & Qwen3-8B\\
\midrule
Epochs & 3 & 3\\
Maximum sequence length & 13,312 & 13,312\\
Prompt / response limits & 12,288 / 1,024 & 12,288 / 1,024\\
OPD learning rate & $10^{-6}$, constant & $10^{-6}$, constant\\
SFT learning rate & $10^{-5}$, cosine & $10^{-5}$, cosine\\
OPD global batch & 32 prompts & 32 prompts\\
SFT global batch & 256 & 256\\
Student rollouts per prompt & 4 & 4\\
Teacher top-$k$ support & 16 & 16\\
Validation interval & 20 steps & 40 steps\\
Training dtype & fp32 & bf16\\
Test decoding & Greedy & Greedy\\
\bottomrule
\end{tabularx}
\caption{Common training and evaluation configuration. Reverse-KL reduction is
computed in fp32 for both model sizes.}
\label{tab:app-train-config}
\end{table}

Thinking is disabled in the chat template, gradient checkpointing and padding
removal are enabled, and evaluation uses greedy decoding with a 1,024-token
response limit. SFT-family runs use AdamW with warmup ratio $0.1$, weight decay
$0.01$, and gradient clipping at $1.0$. Distributional-distillation runs use
constant learning rate, no warmup, no weight decay, temperature $1$, and no
additional gold-CE term.

\subsection{Development choices and checkpoint selection}

Table~\ref{tab:app-development} distinguishes the factors evaluated as reported
method comparisons from settings fixed for the final protocol. The teacher
view, target type, and prefix source form the controlled design analysis in the
main paper. Test-Indist and OOD-600 are evaluated only after checkpoint
selection.

\begin{table*}[t]
\centering
\small
\setlength{\tabcolsep}{5pt}
\renewcommand{\arraystretch}{1.08}
\begin{tabularx}{\textwidth}{@{}>{\raggedright\arraybackslash}p{0.23\textwidth}>{\raggedright\arraybackslash}p{0.40\textwidth}Y@{}}
\toprule
Factor & Values used & Role in the protocol\\
\midrule
Teacher conditioning view
& Original; Oracle State
& controlled design comparison, selected by validation BTA\\
Supervision target
& teacher greedy sequence; token-level soft distribution
& controlled design comparison\\
Prefix source
& gold; teacher; teacher-to-student curriculum; student on-policy
& controlled design comparison\\
SFT loss reduction
& token mean; sequence mean
& reported baseline comparison\\
Training composition
& natural mixture; 2:1 \call{}-to-\answer{} resampling
& reported baseline comparison\\
Objective-specific learning rate
& OPD $10^{-6}$; SFT $10^{-5}$; DPO $5\times10^{-7}$
& fixed by training regime\\
Student rollouts
& four for on-policy training; one for fixed-prefix variants
& fixed by prefix construction\\
Reverse-KL support
& top 16 teacher and student tokens
& fixed protocol setting\\
DPO coefficient
& $\beta=0.1$
& fixed baseline setting\\
Training horizon
& three epochs; one epoch for DPO
& fixed protocol setting\\
\bottomrule
\end{tabularx}
\caption{Development factors, values used, and their role. Method-defining
factors are reported directly as controlled comparisons rather than hidden
hyperparameter selection.}
\label{tab:app-development}
\end{table*}

For every method, all saved checkpoints are scored on the validation split with
\begin{equation}
\mathrm{BTA}_{\mathrm{val}}
=\tfrac{1}{2}\mathrm{TCR}_{\mathrm{val}}
+\tfrac{1}{2}(1-\mathrm{OCR}_{\mathrm{val}}).
\end{equation}
The maximizing checkpoint is frozen before Test-Indist, OOD-600, clean-view,
cross-generator, external-benchmark, and HotpotQA evaluation. Trainers that do
not emit behavioral validation metrics are handled by applying the same
evaluator to each saved checkpoint, preserving one selection rule across all
methods.

\subsection{Final method configurations}

\begin{table*}[t]
\centering
\footnotesize
\setlength{\tabcolsep}{4pt}
\renewcommand{\arraystretch}{1.08}
\begin{tabularx}{\textwidth}{@{}>{\raggedright\arraybackslash}p{0.18\textwidth}>{\raggedright\arraybackslash}p{0.29\textwidth}>{\raggedright\arraybackslash}p{0.25\textwidth}ccc@{}}
\toprule
Method & Training target & Conditioning / prefix & LR & Batch & Epochs\\
\midrule
Gold-SFT
& CE on the source gold action
& Polluted input
& $10^{-5}$ & 256 & 3\\
Sampled-SFT (2:1)
& CE on the source gold action
& Polluted input; 2:1 \call{}:\answer{} sampling
& $10^{-5}$ & 256 & 3\\
SeqNorm-SFT
& sequence-normalized CE
& Polluted input
& $10^{-5}$ & 256 & 3\\
SFT$\rightarrow$DPO
& sigmoid DPO, $\beta=0.1$
& Gold-SFT policy and frozen reference
& $5\times10^{-7}$ & 128 & 1\\
Oracle-SeqKD
& CE on the teacher greedy sequence
& Oracle teacher; fixed sequence target
& $10^{-5}$ & 256 & 3\\
Off-policy OD
& token-level reverse KL
& Oracle teacher; gold prefix
& $10^{-6}$ & 32 & 3\\
Teacher-prefix OD
& token-level reverse KL
& Oracle teacher; teacher prefix
& $10^{-6}$ & 32 & 3\\
Curriculum OPD
& token-level reverse KL
& Oracle teacher; teacher-to-student prefix schedule
& $10^{-6}$ & 32 & 3\\
\best{\ours}
& token-level reverse KL
& Oracle teacher; student on-policy prefix
& $10^{-6}$ & 32 & 3\\
\bottomrule
\end{tabularx}
\caption{Final method-specific configurations. SFT-family learning rates use a
cosine schedule; distillation uses a constant schedule. Batch denotes examples
for SFT/DPO and prompts for distillation.}
\label{tab:app-method-config}
\end{table*}

All distillation variants compute reverse KL over the union of the teacher and
student top-16 supports, use temperature $1$, and set the gold-CE coefficient to
zero. On-policy configurations draw four student continuations per prompt;
fixed-prefix variants use one prefix. The same objective and checkpoint rule are
used for 1.7B$\leftarrow$1.7B, 1.7B$\leftarrow$8B,
8B$\leftarrow$8B, and 8B$\leftarrow$Oracle-SFT-8B transfer. The 1.7B student
is trained in fp32, the 8B student in bf16, and reverse-KL reduction is computed
in fp32 in both cases.

For SFT$\rightarrow$DPO, the preference prompt is the Polluted input, the
preferred continuation is the gold continuation, and the alternative is a
mined tool-use error. Pair construction uses three samples per prompt at
temperature $0.9$; DPO uses per-device batch 2 with gradient accumulation 8 on
eight GPUs.

\subsection{Evaluation protocol and uncertainty}

Each reported cell uses a checkpoint trained under the fixed configuration
recorded in the submitted launch files, followed by deterministic greedy
evaluation. The construction, rollout, resampling, preference-mining,
and bootstrap procedures use fixed seeds recorded in the corresponding
configuration files.

Statistical comparisons use paired resampling over matched evaluation examples.
The primary Test-Indist BTA comparison uses $10{,}000$ paired bootstrap
resamples, and the HotpotQA Noisy-F1 analysis uses $2{,}000$ question-level
paired bootstrap resamples. These analyses quantify evaluation uncertainty for
the frozen checkpoints and do not affect checkpoint selection.

\subsection{Execution environment}

\begin{table*}[t]
\centering
\small
\setlength{\tabcolsep}{6pt}
\renewcommand{\arraystretch}{1.08}
\begin{tabularx}{\textwidth}{@{}>{\raggedright\arraybackslash}p{0.24\textwidth}Y@{}}
\toprule
Component & Configuration\\
\midrule
Training hardware
& one node with $8\times$ NVIDIA H20 GPUs, 96\,GB HBM per GPU\\
Operating environment
& TencentOS Server 4 series, Linux 6.6 series, Python 3.11\\
Core numerical stack
& PyTorch/CUDA with full-parameter FSDP; resolved versions are recorded in the submitted environment manifest\\
Training framework
& verl 0.7.0 with Ray-based distributed training for distillation and torchrun FSDP for SFT\\
Rollout and evaluation
& vLLM with eager execution; tensor parallelism 1 for 1.7B and 8 for 8B evaluation\\
Preference optimization
& TRL \codeid{DPOTrainer}; resolved version recorded in the submitted environment manifest\\
Attention and model stack
& FlashAttention and HuggingFace Transformers; resolved versions recorded in the submitted environment manifest\\
Precision
& 1.7B training in fp32; 8B training in bf16; reverse-KL reduction in fp32\\
\bottomrule
\end{tabularx}
\caption{Hardware and principal software environment. The submitted environment
manifest records the resolved package set used for execution.}
\label{tab:app-environment}
\end{table*}

\subsection{Submitted code and data supplement}

The submitted package maps each paper component to an executable entry point.
Benchmark construction is provided by the five numbered stages,
\codeid{run\_pipeline.sh}, and the shared construction, operator, Oracle, and
LLM-client modules. Method-specific launchers cover SFT, DPO, sequence
knowledge distillation, Oracle-guided distributional distillation, and the
teacher--student scaling runs. The evaluation package contains the shared
behavioral evaluator, validation-BTA checkpoint selector, cross-view and
external-benchmark runners, and paired-bootstrap analysis. The README maps each
reported experiment family to its entry point and records the expected inputs,
outputs, prompt templates, and configuration files.

The submitted data artifact contains the four synchronized benchmark splits,
prompt templates, audit metadata, and record schemas. The third-party APIGen-MT
source is obtained from its original release and is not redistributed.

\subsection{Training and deployment cost}

On eight H20 GPUs, the matched 1.7B Off-policy OD and \ours{} runs require
13.7 and approximately 60 aggregate GPU-hours, respectively. Both deploy the
same compact 1.7B student with identical inference requirements; neither the
teacher nor Oracle State is used after training. With an 8B teacher, the same
deployment model reaches $0.9193$ BTA.

\section{Per-View Evaluation Details}
\label{app:views}

Table~\ref{tab:app-view-decomposition} reports all component metrics for the
frozen checkpoints used in the main clean-transfer comparison. Polluted and
Original inputs share the same current request, tools, and gold action; only
the interaction history differs. Oracle State is used as the diagnostic and
teacher-conditioning view, with the base-model Oracle diagnosis reported in
the main paper. OOD-600 is call-only, and its Polluted/Original TCR appears in
the main clean-transfer table.

\subsection{Polluted- and Original-view metric decomposition}

\begin{table*}[t]
\centering
\scriptsize
\setlength{\tabcolsep}{2.2pt}
\renewcommand{\arraystretch}{1.06}
\begin{tabular}{@{}cclrrrrrr@{}}
\toprule
Scale & View & Method & BTA & TCR & $1-\mathrm{OCR}$ & Dec. & Tool & Arg.$\mid$Tool\\
\midrule
\multirow{8}{*}{1.7B}
& \multirow{4}{*}{P}
& Gold-SFT       & 0.6628 & 0.3422 & \best{0.9834} & 0.6838 & 0.5370 & 0.6373\\
&& Oracle-SeqKD  & 0.8229 & 0.7688 & 0.8771 & 0.9410 & 0.9066 & 0.8481\\
&& Off-policy OD & 0.8499 & 0.9041 & 0.7956 & 0.9376 & 0.9642 & \best{0.9377}\\
&& \best{\ours}  & \best{0.8703} & \best{0.9132} & 0.8273 & \best{0.9473} & \best{0.9824} & 0.9296\\
\cmidrule(l){2-9}
& \multirow{4}{*}{O}
& Gold-SFT       & 0.5880 & 0.1857 & \best{0.9903} & 0.4709 & 0.2409 & 0.7708\\
&& Oracle-SeqKD  & 0.8082 & 0.7876 & 0.8287 & 0.9254 & 0.9059 & 0.8694\\
&& Off-policy OD & 0.8205 & 0.8786 & 0.7624 & 0.9275 & 0.9466 & \best{0.9282}\\
&& \best{\ours}  & \best{0.8450} & \best{0.8890} & 0.8011 & \best{0.9368} & \best{0.9727} & 0.9139\\
\midrule
\multirow{6}{*}{8B}
& \multirow{3}{*}{P}
& Gold-SFT                & 0.9221 & 0.8786 & \best{0.9655} & 0.9368 & 0.9211 & 0.9539\\
&& \ours{} (base teacher)       & 0.9077 & 0.8817 & 0.9337 & 0.9528 & 0.9472 & 0.9308\\
&& \best{\ours{} (Oracle-SFT teacher)} & \best{0.9301} & \best{0.9417} & 0.9185 & \best{0.9604} & \best{0.9763} & \best{0.9646}\\
\cmidrule(l){2-9}
& \multirow{3}{*}{O}
& Gold-SFT                & 0.9086 & 0.8586 & \best{0.9586} & 0.9304 & 0.9144 & 0.9390\\
&& \ours{} (base teacher)       & 0.9092 & 0.8944 & 0.9240 & 0.9507 & 0.9533 & 0.9383\\
&& \best{\ours{} (Oracle-SFT teacher)} & \best{0.9260} & \best{0.9320} & 0.9199 & \best{0.9574} & \best{0.9721} & \best{0.9588}\\
\bottomrule
\end{tabular}
\caption{Test-Indist metric decomposition for the post-trained checkpoints
used in the clean-transfer comparison. P and O denote Polluted and Original
histories. TCR requires a complete correct call; Tool and
Arg.$\mid$Tool denote tool accuracy and argument exact match conditional on
correct tool selection.}
\label{tab:app-view-decomposition}
\end{table*}

Across both scales, the reliable-state configurations retain their advantage
when the misleading traces are removed. The 1.7B student obtains
$0.8703/0.8450$ BTA on Polluted/Original input, while the Oracle-specialized 8B
configuration obtains $0.9301/0.9260$. The decomposition shows that
these gains combine strong complete-call execution with selective non-call
control rather than relying on a view-specific correction rule.

\section{Cross-Generator Evaluation}
\label{app:cross-generator}

To test whether the conclusions depend on the context generator, we construct
an independently generated cross-generator evaluation set with
Claude-Opus-4.7. The set contains 624 synchronized examples, including 412
\call{} and 212 \answer{} decisions. Claude independently generates the
Polluted traces and Oracle States while preserving the source trajectory,
current tools, latest request, and gold next action. All models are evaluated
on the same 624 examples. Checkpoints remain fixed from the original validation
protocol, and no Claude-generated example is used for training, adaptation, or
model selection.

\begin{table*}[t]
\centering
\footnotesize
\setlength{\tabcolsep}{4pt}
\renewcommand{\arraystretch}{1.06}
\begin{tabular}{@{}cllrrr@{}}
\toprule
Scale & Evaluation & Input / method & BTA $\uparrow$ & TCR $\uparrow$ & \shortstack{$1-\mathrm{OCR}$\\$\uparrow$}\\
\midrule
\multirow{6}{*}{1.7B}
& \multirow{3}{*}{Three-view}
& $M_0$ Polluted      & 0.4895 & 0.5922 & 0.3868\\
&& $M_0$ Original     & 0.6454 & 0.7767 & 0.5142\\
&& $M_0$ Oracle State & 0.7557 & 0.9029 & 0.6085\\
\cmidrule(l){2-6}
& \multirow{3}{*}{Polluted transfer}
& Oracle-SeqKD         & 0.7614 & 0.7257 & \best{0.7972}\\
&& Off-policy OD       & 0.7922 & 0.9005 & 0.6840\\
&& \best{\ours}       & \best{0.8135} & \best{0.9053} & 0.7217\\
\midrule
\multirow{6}{*}{8B}
& \multirow{3}{*}{Three-view}
& $M_0$ Polluted      & 0.6867 & 0.8592 & 0.5142\\
&& $M_0$ Original     & 0.8360 & 0.9126 & 0.7594\\
&& $M_0$ Oracle State & 0.8976 & 0.9272 & 0.8679\\
\cmidrule(l){2-6}
& \multirow{3}{*}{Polluted transfer}
& Gold-SFT                         & 0.9120 & 0.8665 & \best{0.9575}\\
&& \ours{} (base teacher)          & 0.8922 & 0.8786 & 0.9057\\
&& \best{\ours{} (Oracle-SFT teacher)}
                                      & \best{0.9238} & \best{0.9466} & 0.9009\\
\bottomrule
\end{tabular}
\caption{Cross-generator evaluation on 624 synchronized contexts independently
generated by Claude-Opus-4.7, comprising 412 \call{} and 212 \answer{}
decisions. The table reports BTA and its complete-call and non-call components.}
\label{tab:app-cross-generator}
\end{table*}

Across both model scales, the base model preserves the reliability ordering
Polluted $<$ Original $<$ Oracle State. Under Polluted input, \ours{} achieves
the strongest 1.7B balance at $0.8135$ BTA, while the Oracle-SFT-teacher
configuration reaches the strongest 8B result at $0.9238$ BTA. The same
diagnosis and method advantage therefore hold under independently generated
contexts.

\section{HotpotQA Noise-Robustness\\Evaluation}
\label{app:hotpotqa}

We evaluate fixed 1.7B deployment checkpoints zero-shot on all 7,405
examples in the HotpotQA-Distractor validation split, varying teacher capacity
while keeping the deployed student fixed. The model predicts only the final
short answer; no supporting-fact supervision or task-specific adaptation is
used. Each item has two paired inputs: \emph{Noisy}, containing the question
and all ten supplied paragraphs, and \emph{Clean}, containing the complete
paragraphs whose titles match the annotated supporting facts. All checkpoints
use the same prompt, temperature 0, a 64-token generation limit, and Qwen3
thinking disabled. Outputs are scored with the official HotpotQA answer
normalization, exact match, and token-level F1. We define the noise gap as
$\mathrm{F1}_{\mathrm{Clean}}-\mathrm{F1}_{\mathrm{Noisy}}$, where a smaller
value indicates stronger retention under distractor context.

\begin{table}[!t]
\centering
\scriptsize
\setlength{\tabcolsep}{1.5pt}
\renewcommand{\arraystretch}{1.04}
\begin{tabularx}{\columnwidth}{@{}>{\raggedright\arraybackslash}Xrrrrr@{}}
\toprule
Method
& \shortstack{Noisy\\EM}
& \shortstack{Noisy\\F1}
& \shortstack{Clean\\EM}
& \shortstack{Clean\\F1}
& Gap $\downarrow$\\
\midrule
Gold-SFT & 0.2941 & 0.4180 & 0.5013 & 0.6398 & 0.2218\\
Off-policy OD & 0.3444 & 0.4715 & 0.5003 & 0.6482 & 0.1767\\
\ours & 0.3457 & 0.4718 & 0.5009 & 0.6483 & 0.1766\\
\best{\ours{} (8B teacher)}
& \best{0.3726} & \best{0.5015}$^{\dagger}$
& \best{0.5095} & \best{0.6611} & \best{0.1596}\\
\bottomrule
\end{tabularx}
\caption{HotpotQA-Distractor results for the compact 1.7B deployment model.}
\label{tab:app-hotpotqa}
\end{table}

For the compact 1.7B deployment model, the 8B-teacher configuration attains
the highest Noisy and Clean F1 and the smallest noise gap. Its $+0.0297$
Noisy-F1 gain over the next-best trained checkpoint is supported by paired
bootstrap over the common $7{,}405$ questions ($95\%$ CI
$[+0.0246,+0.0348]$); relative to Gold-SFT, the gain is $+0.0835$ ($95\%$ CI
$[+0.0754,+0.0913]$). Stronger reliable-state supervision therefore transfers
distractor robustness without increasing deployment size.

\end{document}